\documentclass{article} 
\usepackage{collas2023_conference,times}
\usepackage{easyReview}
\usepackage{dsfont}
\usepackage[noend]{algpseudocode}
\usepackage{algorithm}
\algdef{SE}{Repeat}{EndRepeat}[1]{\textbf{repeat} \(\mbox{#1}\)}{\textbf{end}}%
\usepackage{wrapfig}

\usepackage{amsmath,amsfonts,bm}

\def\eqref#1{equation~\ref{#1}}

\def\1{\bm{1}}

\DeclareMathAlphabet{\mathsfit}{\encodingdefault}{\sfdefault}{m}{sl}
\SetMathAlphabet{\mathsfit}{bold}{\encodingdefault}{\sfdefault}{bx}{n}

\newcommand{\state}{x}
\newcommand{\states}{\mathcal{X}}
\newcommand{\object}{o}
\newcommand{\objects}{\mathcal{O}}\newcommand{\action}{u}
\newcommand{\actions}{\mathcal{U}}
\newcommand{\simulator}{f}
\newcommand{\goal}{g}
\newcommand{\plan}{\pi}
\newcommand{\predicate}{\psi}
\newcommand{\predicates}{\Psi}
\newcommand{\classifier}{c}
\newcommand{\class}{\ell}
\newcommand{\network}{h}
\newcommand{\type}{\lambda}
\newcommand{\types}{\Lambda}
\newcommand{\ground}{\underline}
\newcommand{\dataset}{\mathcal{D}}
\newcommand{\horizon}{h}
\newcommand{\query}{\mathcal{Q}}
\newcommand{\abstractstate}{s}
\newcommand{\abstractfn}{\texttt{abstract}}
\newcommand{\abstractaction}{a}
\newcommand{\abstractactions}{\mathcal{A}}
\newcommand{\abstracttransitionfn}{F}
\newcommand{\querypolicy}{\pi_{\text{query}}}
\newcommand{\samplers}{\Omega}

\usepackage{hyperref}
\hypersetup{
    colorlinks=true,
    linkcolor=red,
    filecolor=magenta,
    urlcolor=blue,
    citecolor=purple,
    pdftitle={Embodied Active Learning of Relational State Abstractions for Bilevel Planning},
    pdfpagemode=FullScreen,
    }

\title{Embodied Active Learning of Relational State Abstractions for Bilevel Planning}

\author{Amber Li  \\
MIT CSAIL \\
\texttt{amli@alum.mit.edu} \\
\And 
Tom Silver  \\
MIT CSAIL \\
\texttt{tslvr@mit.edu} \\
}

\collasfinalcopy 

\begin{document}

\maketitle

\begin{abstract}
State abstraction is an effective technique for planning in robotics environments with continuous states and actions, long task horizons, and sparse feedback.
In object-oriented environments, predicates are a particularly useful form of state abstraction because of their compatibility with symbolic planners and their capacity for relational generalization.
However, to plan with predicates, the agent must be able to interpret them in continuous environment states (i.e., ground the symbols).
Manually programming predicate interpretations can be difficult, so we would instead like to learn them from data.
We propose an embodied active learning paradigm where the agent learns predicate interpretations through online interaction with an expert.
For example, after taking actions in a block stacking environment, the agent may ask the expert: ``Is On(block1, block2) true?''
From this experience, the agent \emph{learns to plan}: it learns neural predicate interpretations, symbolic planning operators, and neural samplers that can be used for bilevel planning.
During exploration, the agent \emph{plans to learn}: it uses its current models to select actions towards generating informative expert queries.
We learn predicate interpretations as ensembles of neural networks and use their entropy to measure the informativeness of potential queries.
We evaluate this approach in three robotic environments and find that it consistently outperforms six baselines while exhibiting sample efficiency in two key metrics: number of environment interactions, and number of queries to the expert.
\end{abstract}

\section{Introduction}
\label{sec:intro}

\begin{figure}[b]
    \centering
	\includegraphics[width=0.9\textwidth]{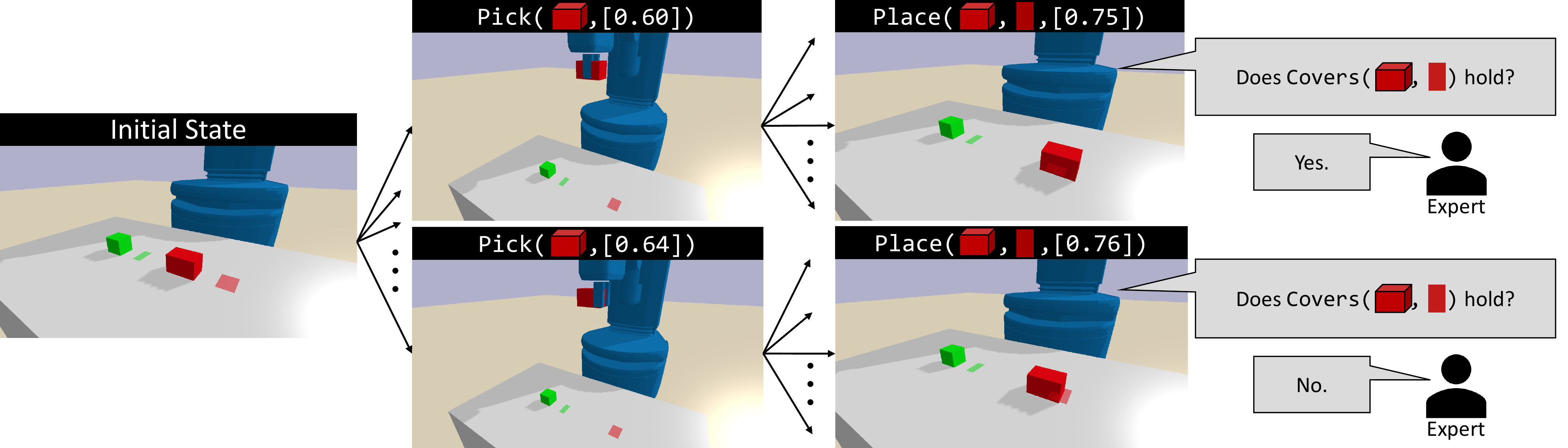}
	\caption{\textbf{Active predicate learning in the PickPlace1D environment}. The \texttt{Covers} predicate is difficult to interpret: given limited data, the agent may not know how to classify a block partially overlapping a region. To improve its interpretations, the agent must take actions to set up classification problems and then query the expert for labels. The figure shows two possible trajectories among infinitely many. There are also queries at intermediate states (not shown).}
  \label{fig:example_states}
\end{figure}

Our research objective is to develop a robotic agent that can achieve a wide variety of high-level goals, like preparing a meal or cleaning up a kitchen, in environments with continuous state and action spaces, long task horizons, and complex constraints.
In this work, we study an \emph{embodied active learning} paradigm, where the robot learns by interacting with its environment, querying expert knowledge, and using the expert's feedback to guide its subsequent exploration and queries \citep{daniel2014active}. Since real-world exploration and data collection is expensive, we want the robot to 1) minimize the number of actions taken in the environment and 2) ask the expert as few questions as possible. In other words, the agent must select actions and query strategically.

Towards achieving a wide distribution of goals in robotic environments, we consider an agent that is \emph{learning to plan}.
In particular, we build on \emph{task and motion planning (TAMP)} \citep{garrett2020integrated}, which uses state and action abstractions to plan efficiently in continuous environments.
Previous work has shown how to learn action abstractions (operators and samplers) when \emph{given} state abstractions (predicates) for TAMP~\citep{silver2021learning,chitnis2021learning}.
However, hand-specifying the state abstractions can be tedious and impractical, even for an expert programmer.
In this work, we consider the problem of \emph{learning} these state abstractions via embodied active learning.
State abstractions in TAMP take the form of \emph{predicates}.
A predicate is a named relation over objects, and the semantic interpretation of a predicate is defined by a binary classifier.
For example, in the PickPlace1D environment~\citep{silver2021learning,chitnis2021learning} (Figure \ref{fig:example_states}), a predicate called \texttt{Covers} takes two object arguments, a \texttt{block} and a \texttt{target}, and the associated classifier returns true if the block completely covers the target.
Applying a set of predicate classifiers to a continuous state induces a discrete abstract state, e.g., \{\texttt{Covers(b1, t1)}, \texttt{HandEmpty(rob)}, \dots\}.
Given a predicate-based goal, TAMP searches in the abstract state space to constrain search in the continuous state space.

We propose \emph{active predicate learning for TAMP}.
A robot is situated in a deterministic environment with an expert.
To begin, the expert gives a small number of demonstrations (to illustrate the task distribution of interest) and a very small number of classification examples (one positive and one negative) for each predicate.
At this point, the robot knows the predicates but not their interpretations; in other words, it needs to solve a symbol grounding problem \citep{harnad1990symbol}\footnote{Another aspect of symbol grounding, which we do not address here, is generating referents for objects.}.
The robot starts to \emph{explore} its environment:
at each step, the robot selects an \emph{action} to execute and a \emph{query} to give the expert.
The query is a set of zero or more ground atoms (predicates with object arguments) that the robot wants to ``check'' in the current state.
For example, querying \{\texttt{Covers(b1, t1)}\} would ask if \texttt{b1} currently covers \texttt{t1} according to the expert's interpretation.
The expert answers ``yes'' or ``no'' according to a noise-free but unknown ground-truth interpretation.
To deal with possible dead-ends, the expert also periodically resets the environment to an initial state drawn from a distribution.
This setting is reminiscent of how a young child might use very sparse linguistic labels in early concept learning~\citep{bowerman2001language,casasola2007novel}.
To measure the extent to which the robot uses its experience to improve its planning ability, we evaluate the robot on a set of held-out planning tasks.

In this setting, the agent is faced with two interrelated subproblems: how to query, and how to select actions.
For example, towards learning the meaning of \texttt{Covers}, querying about a block that partially overlaps a target may be more informative than querying about a block that is far from a target.
Furthermore, the agent may need to carefully select a grasp and place position to reach an ``interesting'' state where there is partial overlap to ask about (Figure \ref{fig:example_states}).
This need for action selection is what distinguishes embodied active learning from typical active learning~\citep{settles2011theories}, and
the availability of an expert to query distinguishes the setting from exploration in (model-based) reinforcement learning~\citep{kaelbling1996reinforcement}.
Nonetheless, we can draw on both of these lines of work to make progress here.

We propose an \emph{action selection strategy} and a \emph{query policy} for active predicate learning.
Both are rooted in the active learning principle that the robot should reduce its uncertainty about its classifiers.
The query policy selects ground atoms whose classification entropy is above a certain threshold.
Action selection uses the robot's learned predicates, operators, and samplers to \emph{plan} to reach states where there is high entropy.
In experiments, we compare against alternative action selection and query policies and find that our main approach effectively balances action cost (number of environment transitions) and query cost (number of ground atoms asked).
In summary, we (1) propose the problem setting of active predicate learning for TAMP; (2) propose an entropy-based, model-based approach; and (3) evaluate the approach in simulated robotic environments.

\section{Problem Setting}\label{sec:problem-setting}

\emph{Environments.}
We consider a robot exploring an environment with deterministic transitions and fully-observed states.
A state $\state \in \states$ is defined by a set of objects $\objects$ and a real-valued feature vector for each object.
The dimensionality of an object's feature vector is determined by the object's \emph{type} $\type \in \types$.
For example, an object of type \texttt{block} may have a feature vector of dimension 4 describing its current pose (x, y, and z coordinates) and color.
An action $\action \in \actions$ is a controller  with discrete and continuous parameters.
For example, $\texttt{Pick(b1,[0.3,0.2,0.4])}$ is an action for picking block $\texttt{b1}$ with continuous grasp pose $\texttt{[0.3,0.2,0.4]}$.
A deterministic simulator $\simulator: \states \times \actions \to \states$ predicts a next state given a current state and action.
The simulator is known to the robot\footnote{Previous work by \citet{chitnis2021learning} has shown that this simulator can also be learned.}, who can use it to plan.
An environment can be viewed as a form of object-oriented~\citep{diuk2008object} or relational~\citep{guestrin2003generalizing} MDP, but note the continuous states and hybrid discrete-continuous actions.

\emph{Predicates.}
A \emph{predicate} $\predicate$ consists of a name (e.g., \texttt{Covers}) and a tuple of typed placeholders for objects $(\type_1, \dots, \type_m)$ (e.g., \texttt{(?block, ?target)}).
The \emph{interpretation} of a predicate is a classifier $\classifier_\predicate : \states \times \objects^m \to \{\text{true, false}\}$.
These classifiers are unknown to the agent and must be learned.
We distinguish between ground atoms, where a predicate is applied to specific objects $(\object_1, \dots, \object_m)$, and lifted atoms, where the predicate is applied to typed variables.
For example, \texttt{Covers(b1, t1)} is a ground atom and \texttt{Covers(?block, ?target)} is a lifted atom. The interpretation of a ground atom $\ground{\predicate}$ with objects $\overline{\object} = (\object_1, \dots, \object_m)$ is given by $\classifier_{\ground{\predicate}}(\state) := \classifier_\predicate(\state, \overline{\object})$.


\emph{Initialization.}
Before exploration, the robot is presented with a small set of demonstrations.
Each demonstration consists of a \emph{task} and a \emph{plan}.
The task consists of an initial state $\state_0 \in \states$ and a goal $\goal$.
The goal is a set of ground atoms and is said to \emph{hold} in a state $\state$ if $\classifier^*_{\ground{\predicate}}(\state) = \text{true}$ for all ground atoms $\ground{\predicate} \in \goal$, where $\classifier^*_{\ground{\predicate}}$ is the (unknown) expert interpretation of $\ground{\predicate}$.
A plan is a sequence of actions $\plan^* = (\action_1, \dots, \action_n)$.
The plan need not be optimal, but it is assumed to solve the task, i.e., simulating $\plan^*$ forward from $\state_0$ will terminate at a state where $\goal$ holds.
The expert additionally presents a very small\footnote{In experiments, we use one positive and one negative example per predicate.} set of examples $\dataset = \{(\state, \ground{\psi}, \class)\}$, where $\class \in \{\text{true}, \text{false}\}$ is the output of $\ground{\psi}(\state)$ under the expert's interpretation.
This dataset communicates the full set of predicates $\predicates$ that the robot can use to query the expert.
In other words, the expert assumes that the robot will extract the predicates in $\dataset$ -- \texttt{Covers}, \texttt{Holding}, and so on -- and use them to form queries.

\emph{Exploration and evaluation.}
After initialization, the robot begins to explore the environment.
At the start of each \emph{episode} of exploration, the environment is reset to an initial state $\state_0 \in \states$ sampled from an initial state distribution.
Then, for up to $\horizon$ steps, the robot repeatedly queries the expert about the current state $\state$ and executes an action to advance the state $\state'$.
A \emph{query} $\query$ is a set of ground atoms, and a \emph{response} is a set of $(\ground{\predicate}, \class)$ where $\ground{\predicate} \in \query$ and $\class = \classifier^*_{\ground{\predicate}}(\state)$.
An example query in PickPlace1D is \{\texttt{Covers(b1, t1)}, \texttt{Covers(b1, t2)}\}, and a possible response is \{(\texttt{Covers(b1, t1)}, True), (\texttt{Covers(b1, t2)}, False)\}.
Each response is added to the robot's dataset with the current state, i.e., $\dataset \gets \dataset \cup \{(\state, \ground{\psi}, \class)\}$.
To measure progress, we periodically \emph{evaluate} the robot on a set of held-out tasks.
Each task $\langle \state_0, \goal \rangle$ is considered solved if the robot reaches $\goal$ from $\state_0$ within $\horizon$ steps and within a planning timeout.
The robot's objective is to take actions, make queries, and use the responses to maximize the number of tasks solved after a minimal number of exploration actions and queries.

\begin{figure}[t]
    \centering
	\includegraphics[width=0.9\textwidth]{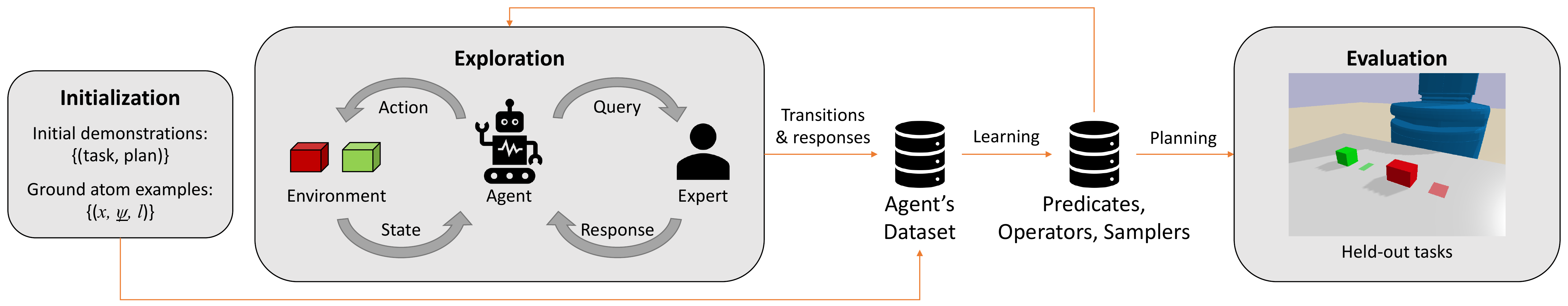}
	\caption{\textbf{Problem setting and approach overview.} The agent's dataset is initialized with a small number of demonstrations and ground atom examples, which it uses to learn initial predicates, operators, and samplers. Those models are then used during exploration, where the agent takes actions in the environment and queries the expert. From these interactions, the dataset grows and the models improve. We periodically evaluate the agent on held-out planning tasks.}
  \label{fig:problem_setting}
\end{figure}

\section{Learning Abstractions for Bilevel Planning}\label{sec:background}

Our work builds on recent advances in learning abstractions for bilevel planning~\citep{silver2021learning,chitnis2021learning}, a specific instantiation of TAMP.
We review the key ideas here and refer readers to the references for details.

\subsection{Bilevel Planning with Predicates, Operators, and Samplers}\label{sec:planning}

\begin{wrapfigure}{r}{0.3\textwidth}
\vspace{-4em}
\includegraphics[width=\linewidth]{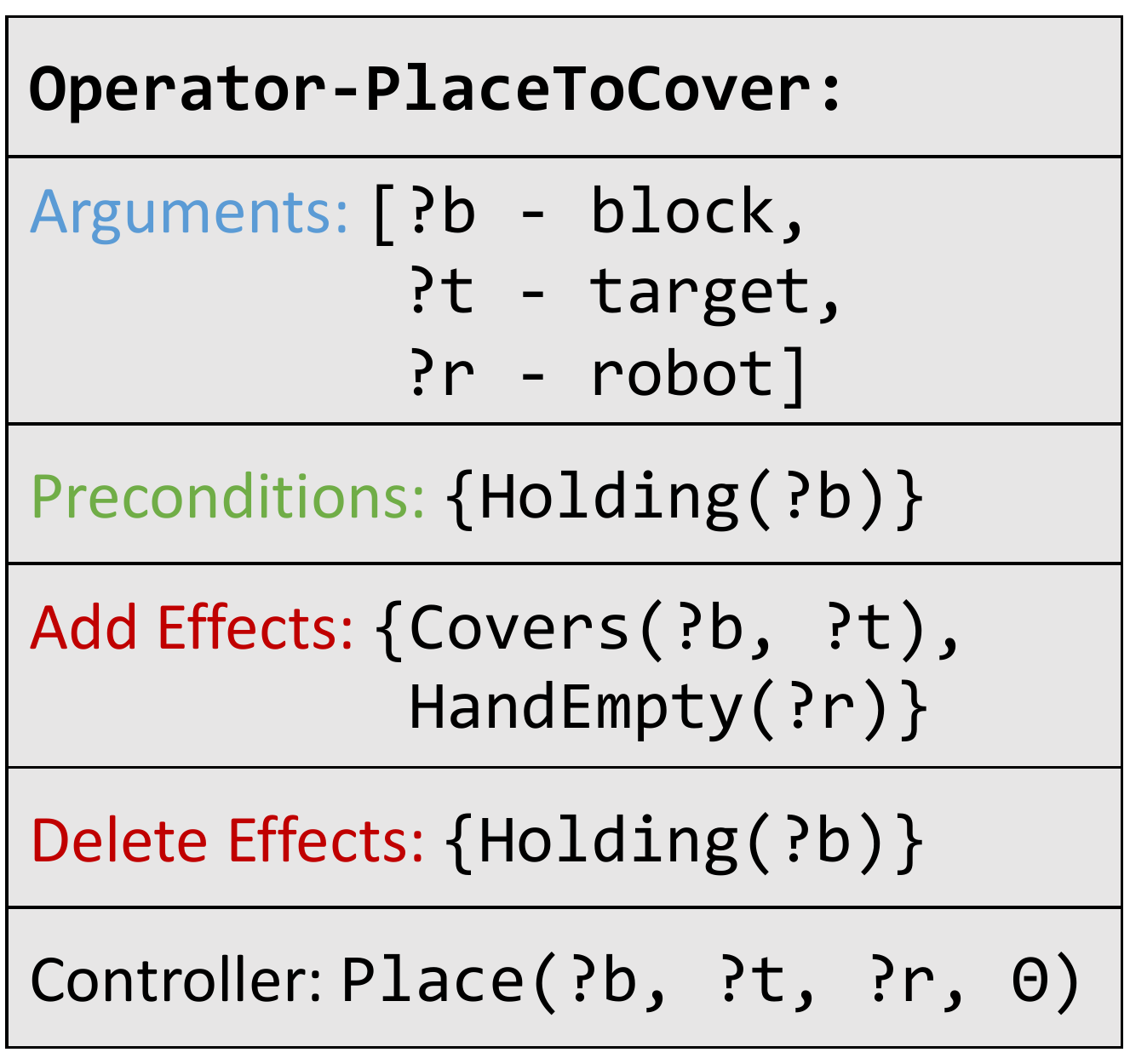} 
\vspace{-3em}
\end{wrapfigure}

The first key idea in bilevel planning is that \emph{predicates induce abstract states}.
In particular, given a state $\state$, a set of predicates $\predicates$, and their interpretations, we can create a corresponding abstract state 
$$\abstractfn(\state, \predicates) := \{\ground{\predicate} : \classifier_{\ground{\predicate}}(\state) = \text{true}\}.$$
We use $\abstractstate$ to denote an abstract state.
An abstract state will generally lose information about the original state.
However, if the predicates are defined judiciously, they can provide guidance for planning.

The second key idea is that \emph{abstract actions define transitions between abstract states.}
Abstract actions are defined in terms of \emph{operators} and \emph{samplers}.
An operator has arguments, preconditions, effects, and a controller.
We eschew formal definitions \citep{silver2021learning,chitnis2021learning} in favor of simplified exposition and refer to the example on the right.
The arguments are variables, i.e., typed placeholders for objects.
The preconditions are lifted atoms that define what must be true in an abstract state for this operator to be applied.
The effects determine how the abstract state would change if this operator were applied; add effects are added, and delete effects are removed.
Finally, the controller connects the abstract action to the environment action space.
The discrete parameters of the controller (e.g, which target to place on) are determined in the operator, but the continuous parameters (e.g., what position offset to use) are undetermined.
To propose different values for the continuous parameters, a \emph{sampler} is associated with the operator.
A \emph{ground operator} is an operator whose arguments have been substituted for objects.
The assignment of arguments to objects is also given to the sampler, along with the current state, so that the sampler can propose targeted values for the controller.
We use $\abstractaction \in \abstractactions$ to denote a ground operator and $\abstracttransitionfn(\abstractstate, \abstractaction) = \abstractstate'$ to denote the (partial) \emph{abstract transition function} induced by the operators.

Given predicates, operators, samplers, and a task $\langle \state_0, \goal \rangle$, bilevel planning generates candidate \emph{abstract plans} and then attempts to \emph{refine} those plans into environment actions~\citep{silver2021learning,chitnis2021learning}.
An abstract plan comprises a \emph{subgoal sequence} and a \emph{plan sketch}.
The subgoal sequence consists of abstract states $(\abstractstate_0, \dots, \abstractstate_n)$ where $\abstractstate_0 = \abstractfn(\state_0, \predicates)$ and $\goal \subseteq \abstractstate_n$.
For example, if $\abstractstate_1 = \{\texttt{Holding(rob, b1)}\}$, then the robot will attempt to find a continuous action that leads to it holding \texttt{b1}.
The plan sketch is a sequence of ground operators $(\abstractaction_1, \dots, \abstractaction_n)$ and their associated samplers.
Abstract plans are generated iteratively using an AI planner~\citep{hoffmann2001ff,helmert2006fast}.
For each abstract plan, the samplers in the plan sketch are repeatedly invoked until all subgoals are reached, or until a maximum number of tries is exceeded, at which point the next abstract plan is considered.

\subsection{Learning Operators and Samplers Given Predicates}\label{sec:learning-ops}

Our focus in this work is on learning predicate interpretations through embodied active learning.
Given predicate interpretations, previous work has shown how to learn operators~\citep{silver2021learning} and samplers~\citep{chitnis2021learning}.
We use these techniques without modification and describe them very briefly here.

Given a dataset of transitions $(\state, \action, \state')$, we can create a corresponding dataset of abstract state transitions $(\abstractstate, \action, \abstractstate')$ where $\abstractstate = \abstractfn(\state, \predicates)$ and $\abstractstate' = \abstractfn(\state', \predicates)$.
To learn operators, the latter dataset is first partitioned so that two transitions are in the same partition set if their controllers and effects (changes in abstract state) are equivalent up to object substitution.
For example, all transitions where a \texttt{Place} controller was used to successfully achieve \texttt{Covers} would be grouped together.
For each partition set, preconditions are determined by finding all atoms in common at the start of each transition, again up to object substitution.
This can be calculated efficiently via set intersection after objects are replaced with variable placeholders, which in turn become arguments.
With arguments, preconditions, effects, and controllers determined for each partition set, the operators are complete.

The partition of abstract transitions is used once more for sampler learning.
Each sampler is implemented as a neural network that takes in the object features of the operator arguments and returns the mean and diagonal covariance of a multivariate Gaussian distribution.
The Gaussian is then used to propose values for the controller parameters.
Training data for each sampler is extracted from the respective partition set and the neural networks are trained to minimize a Gaussian negative log-likelihood loss.

\section{Active Predicate Learning}\label{sec:methods}

We want the robot to explore \emph{efficiently} so that it can maximize performance on our real objective: \emph{effectively} solving held-out planning tasks.
As the robot explores, its dataset of environment transitions and query responses will grow.
How should it make use of these data?
Where and what should it explore next?
We propose that the robot should \emph{learn to plan} and then \emph{plan to explore}.

\subsection{Neural Predicate Learning}

Recall that a dataset $\dataset$ of query responses $(\state, \ground{\predicate}, \class)$ is given to the robot during initialization and then extended during exploration.
We use these data to train \emph{neural predicate classifiers} (interpretations).
Each classifier $\classifier_\predicate$ is parameterized as an ensemble of $k$ fully-connected neural networks $\network^{(1)}_{\classifier_\predicate}, \dots, \network^{(k)}_{\classifier_\predicate}$.
Each member of the ensemble $\network^{(i)}_{\classifier_\predicate}$ maps a state $\state$ and objects $\overline{\object} = (\object_1, \dots, \object_m)$ to a probability that the class is true.
Since the full object set $\objects$ can vary in size between tasks, we make the simplifying assumption that the only objects relevant to a predicate interpretation are those present in the arguments\footnote{In general, this assumption is limiting. Relational neural networks (e.g., GNNs) may be used to avoid this assumption.}.
We then parameterize each ensemble member as
$$\network^{(i)}_{\classifier_\predicate}(\state[\object_1] \oplus \dots \oplus \state[\object_m]),$$
where $\state[\object]$ denotes the feature vector of $\object$ in $\state$ and $\oplus$ denotes concatenation.
The final output of the classifier $\classifier_\predicate(\state)$ is true if the average predicted probability of the ensemble members exceeds 0.5.

Since the predicate arguments are typed and feature dimensions are fixed per type, the input to each ensemble member is a fixed-dimensional vector.
Thus, if we can construct input-output training examples from $\dataset$, we can use standard neural network classifier training techniques.
To construct these examples, we partition $\dataset$ into predicate-specific datasets, where the dataset for predicate $\predicate$ consists of $(\state, \overline{\object}, \class)$ tuples where $(\state, \ground{\predicate}, \class) \in \dataset$ and $\overline{\object} = (\object_1, \dots, \object_m)$ are the objects used to ground $\predicate$ in $\ground{\predicate}$.
For example, if $\predicate = \texttt{Covers}$ and $\ground{\predicate} = \texttt{Covers(b1, t1)}$, then $\overline{\object} = \texttt{(b1, t1)}$.
We further transform each $(\state, \overline{\object}, \class)$ tuple into an input vector $\state[\object_1] \oplus \dots \oplus \state[\object_m]$ and output class $\class$.
With these data, we optimize the weights of $\network^{(i)}_{\classifier_\predicate}$ to minimize binary cross-entropy loss via Adam~\citep{kingma2014adam}.

We use ensembles of neural networks because they provide a measure of uncertainty, which we will later leverage during exploration.
(Other uncertainty quantification strategies are possible.)
To this end, it is important that the networks converge to different hypotheses when there are many possible explanations of limited data.
One way to promote diversity between networks is to initialize their weights differently at the start of training.
In preliminary experiments, we found that increasing the variance of weight initialization led to greater ensemble disagreement but also convergence failures if the variance was too high.
In our main experiments, we initialize network weights via a unit Gaussian distribution, detect convergence failures, and restart training if necessary; see Appendix~\ref{app:experiment-details} for details.

With predicate interpretations learned, we can then apply the techniques from previous work to learn operators and samplers (Section \ref{sec:learning-ops}).
This whole training pipeline is executed for the first time during the initialization phase of active predicate learning (Section \ref{sec:problem-setting}) and repeated after each episode of exploration.
Given predicates, operators, and samplers, we have all the components needed for planning (Section \ref{sec:planning}).
We can use this ability not only to solve held-out problems during evaluation, but also to guide exploration.

\subsection{Model-based Exploration}

Given predicates, operators, and samplers learned from the data collected so far, how should the robot collect more data to improve these models?
To answer this question, we must define mechanisms for (1) query generation and (2) sequential action selection.
In generating queries, the robot should reason about the value of different possible queries and trade off the need to collect more data with the cost of burdening the expert.
In selecting actions, the robot should seek out regions of the state space where it can gather the most information to improve its models.

\subsubsection{Query Generation}

When the robot is in a state and deciding what queries to give the expert, it is solving an \emph{active learning} problem.
One of the main principles in active learning is that queries should be selected on the basis of the robot's \emph{epistemic uncertainty} about potential responses.
For example, if the robot is confident that \texttt{GripperOpen(rob)} is true and \texttt{Holding(b1, rob)} is false in the current state, then neither ground atom would be worth including in a query.
If the robot is more unsure about \texttt{Covers(b1, t1)}, then that ground atom would be a better choice.

We use classifier \emph{entropy} as a measure of epistemic uncertainty.
Let $P(\classifier_{\ground{\predicate}}(\state) = \class) = \frac1k \sum_{i=1}^k P(\network^{i}_{\ground{\predicate}}(\state) = \class)$ denote the probability that the interpretation of ground atom $\ground{\predicate}$ is $\class$ in state $\state$ according the robot's current ensemble.
The entropy for $\ground{\predicate}$ in $\state$ is then
\begin{equation*}
\texttt{entropy}(\ground{\predicate}, \state) :=  - \sum_{\class=0, 1} \left(P\left(\classifier_{\ground{\predicate}}(\state) = \class \right)\right) \log \left( P\left(\classifier_{\ground{\predicate}}(\state) = \class \right)\right).
\end{equation*}
We use entropy to define a \emph{query policy}:
\begin{equation*}
\querypolicy(\state) = \{ \ground{\predicate}: \texttt{entropy}(\ground{\predicate}, \state) > \alpha, \forall \ground{\predicate} \in \ground{\predicates} \},
\end{equation*}
where $\alpha$ is a hyperparameter ($\alpha=0.05$ in experiments) and $\ground{\predicates}$ is the set of all ground atoms.
This query policy dictates that the robot will ask the expert about all ground atoms whose interpretations in the current state are sufficiently uncertain.
The policy is \emph{greedy} in the sense that it only uses the robot's current uncertainty, rather than predicting how its uncertainty would change given different responses~\citep{settles2011theories}.
Nonetheless, as the robot collects more data and revises its predicate interpretations, its uncertainty will generally decrease, and the number of ground atoms included in queries will generally decline.

\subsubsection{Sequential Action Selection}

\begin{wrapfigure}{R}{0.5\textwidth}
\vspace{-1em}
\begin{minipage}{0.5\textwidth}
\begin{algorithm}[H]
\caption{Lookahead Action Selection}
\label{alg:lookahead}
\begin{algorithmic}[1]
\State \textbf{inputs:} state $\state_0$, predicates $\predicates$, ground operators $\abstractactions$,
\State {\color{white}\textbf{Inputs:}} learned samplers $\samplers$, simulator $\simulator$
\State \textbf{hyperparameters:} \texttt{maxTrajs}, \texttt{maxHorizon}
\Repeat{\texttt{maxTrajs} times}
\State $\state \gets \state_0$
\State $\texttt{score} \gets 0$
\Repeat{\texttt{maxHorizon} times}
\State $\abstractstate \gets \abstractfn(\state, \predicates)$
\State $\abstractaction \gets \texttt{sampleApplicableOp}(\abstractstate, \abstractactions)$
\State $\action \gets \texttt{sampleAction}(\abstractaction, \state, \samplers)$
\State $\state \gets \simulator(\state, \action)$
\State $\texttt{stateScore} \gets \sum_{\ground{\predicate}}\texttt{entropy}(\ground{\predicate}, \state)$
\State $\texttt{score} \gets \texttt{score} + \texttt{stateScore}$
\EndRepeat
\EndRepeat
\State \Return action trajectory that maximized \texttt{score}
\end{algorithmic}
\end{algorithm}
\end{minipage}
\vspace{-1em}
\end{wrapfigure}

After the robot generates a query and receives a response, it must select an action to take.
In practice, the robot will make an entire plan and then generate a query at each of the states it encounters.
The main consideration in action selection is that the robot should visit states that allow for informative queries.
For example, in the PickPlace1D environment (Figure \ref{fig:example_states}), the robot may be very uncertain about the interpretation of \texttt{Covers} in the case where a block is overlapping, but not completely covering, a target region.
Since blocks are always disjoint from targets in initial states, the robot would need to carefully select a \texttt{Pick} and a \texttt{Place} action before it can ask the expert about this case.

Since the robot is learning models for bilevel planning, a natural question is whether we can leverage these models for action selection during exploration.
Previous work on \emph{Goal-Literal Babbling (GLIB)} has shown that planning to achieve randomly sampled goals can be an effective strategy for online operator learning in the case where predicates are known~\citep{chitnis2020glib}.
However, since goals are discrete atoms, GLIB is unable to pursue specific low-level states.
For example, even if GLIB sampled a goal with \texttt{Covers(b1, t1)}, the bilevel planner described in Section~\ref{sec:planning} would have no mechanism to seek out an information-rich state where \texttt{b1} is partially overlapping \texttt{t1}.
Thus, this new problem setting where we are learning not only operators but also predicates (and samplers) through online interaction calls for a different action selection strategy.

We propose a \emph{lookahead} action selection strategy that uses the robot's current models for planning while taking into account the information value of candidate low-level states.
The strategy is summarized in Algorithm~\ref{alg:lookahead}.
Given an initial state and the robot's current predicates, operators, and samplers, the robot samples and simulates \texttt{maxTrajs} possible trajectories.
Each trajectory is sampled by repeatedly abstracting the state using the learned predicate interpretations (Line 8), sampling a learned operator whose preconditions hold (Line 9), sampling an action using the learned samplers (Line 10), and advancing the state (Line 11).
Each state encountered is scored according to the total entropy over all ground atoms (Line 12), and each trajectory is scored by accumulating these scores over all encountered states.
Finally, the actions from the trajectory with the highest score are selected for execution in the environment (Line 16).
In practice, in the case where no applicable operators can be found, we terminate the trajectory early.
Furthermore, in the case where no nontrivial trajectory can be found, we fall back to sampling a random action \citep{chitnis2020glib}.

This lookahead action selection strategy is closely tied to query generation: the robot will seek out states with high entropy, and then query the expert to reduce its uncertainty in those states.
We hypothesize that this tight relationship is essential for efficient active predicate learning.
Furthermore, we hypothesize that exploring to reduce predicate entropy will sufficiently drive operator and sampler learning\footnote{However, in cases where the predicate interpretations are easy to learn, or already known, additional exploration techniques (e.g., GLIB~\citep{chitnis2020glib}) may be useful to drive operator and sampler learning.}.
To test these hypotheses, we turn to experiments.

\section{Experiments}

We now present experimental results evaluating the extent to which our main approach effectively and efficiently learns predicates useful for bilevel planning.
We evaluate seven approaches (main and six baselines) in three environments.

\subsection{Experimental Setup}

\paragraph{Approaches.} Here we briefly describe the approaches, with additional details in Appendix~\ref{app:experiment-details}.
In addition to our main approach, we consider three {\color{black} action selection baselines} and three {\color{black} query generation baselines}.

\begin{itemize}
    \item \textbf{Main.} Our main approach, which uses lookahead action selection and the entropy-based query policy.
    
    \item {\color{black} \textbf{GLIB.}} Same as Main except GLIB~\citep{chitnis2020glib} is used for action selection instead of lookahead.
    
    \item {\color{black} \textbf{Random Actions.}} Same as Main except actions are selected uniformly at random.
    
    \item {\color{black} \textbf{No Actions.}} Same as Main except no actions are taken during exploration (only initial states are queried).
    
    \item {\color{black} \textbf{Ask All.}} Same as Main except all possible queries are generated at every step of exploration.
    
    \item {\color{black} \textbf{Ask None.}} Same as Main except no queries are generated.
    
    \item {\color{black} \textbf{Ask Randomly.}} Same as Main except queries are selected uniformly at random from the set of all possible ground atoms. The number of ground atoms in the query is approximately equal to the number generated by the main entropy-based query policy.
\end{itemize}

\paragraph{Environments.}

We now briefly describe the environments, with additional details in Appendix~\ref{app:experiment-details}.

\begin{itemize}
    \item \textbf{PickPlace1D.} As described in Section \ref{sec:intro}, this environment features a robot that must pick blocks and place them to completely cover target regions along a table surface. All pick and place poses are in a 1D line. Evaluation tasks require 1--4 actions to solve. The predicates are \texttt{Covers}, \texttt{Holding}, and \texttt{HandEmpty}. This environment was proposed by~\citet{silver2021learning}, who used manually designed predicates.
    \item \textbf{Two Rooms.} This is a novel environment that is very loosely inspired by the continuous playroom of \citet{konidaris2009efficient}. Two rooms are connected by a hallway. One room has a table with 3 blocks; the other room has a continuous dial for turning a light on or off. Evaluation tasks require 4--8 actions to solve. The predicates are \texttt{On}, \texttt{OnTable}, \texttt{GripperOpen}, \texttt{Holding}, \texttt{Clear}, \texttt{NextToTable}, \texttt{NextToDial}, \texttt{LightOn}, \texttt{LightOff}.
    \item \textbf{Blocks.} This is a robotic version of the classic blocks world environment. During exploration, 3 or 4 blocks are available. Evaluation tasks have 5 or 6 blocks and require 2--20 actions to solve. The predicates are \texttt{On}, \texttt{OnTable}, \texttt{GripperOpen}, \texttt{Holding}, \texttt{Clear}. This environment was also used by~\citet{silver2021learning} with manually designed predicates.
\end{itemize}

\paragraph{Experimental details.}
All approaches are run across all environments for 1000 transitions and evaluated after every episode on 50 held-out evaluation tasks.
Each trial is repeated over 10 random seeds.
Our key metrics are (1) number of evaluation tasks solved within a planning timeout (10 seconds) and (2) cumulative query cost (total number of ground atoms asked).
Exploration episode lengths are 3, 8, and 20 steps for PickPlace1D, Two Rooms, and Blocks respectively.
Demonstrations in the initial dataset are generated with environment-specific scripts (50 per environment).
Query responses are generated automatically via scripted predicate interpretations.
The initial dataset includes 1 positive and 1 negative example, selected randomly, of each predicate in each environment.
All experiments were conducted on a quad-core Intel Xeon Platinum 8260 processor.
See Appendix~\ref{app:experiment-details} for additional experimental details.

\subsection{Results \& Discussion}

\begin{figure}[t]
    \centering
	\includegraphics[width=0.9\textwidth]{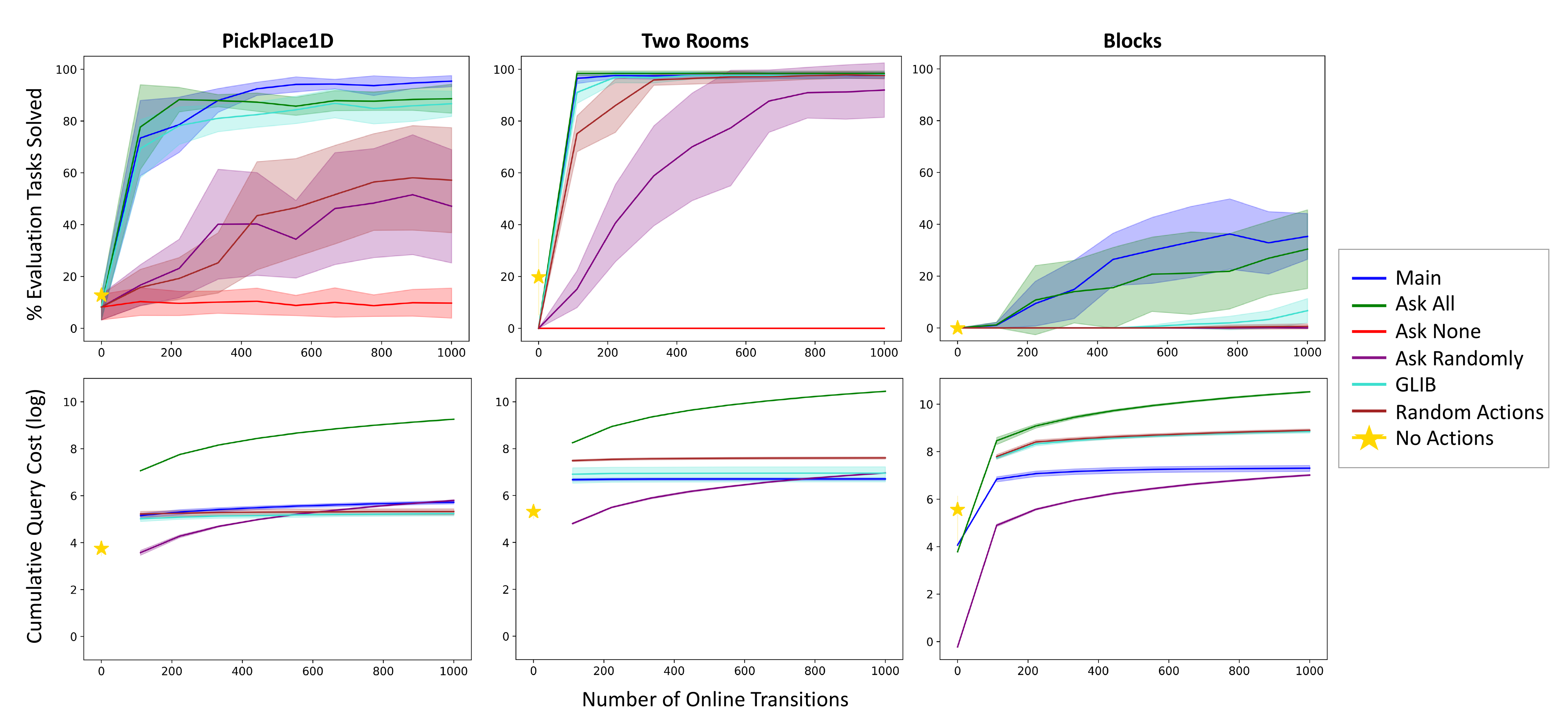}
	\caption[]{\textbf{Main results.} Our main approach solves the same number of evaluation tasks as Ask All (top) using far fewer expert queries (bottom). Query cost is the total number of ground atoms included in queries over time. All results are averaged over 10 random seeds. Lines are means and shaded regions are 95\% $t$-confidence intervals\footnotemark. Note that No Actions is a single point at 0.}
  \label{fig:main_results}
\end{figure}

\begin{figure}[t]
    \centering
	\includegraphics[width=0.9\textwidth]{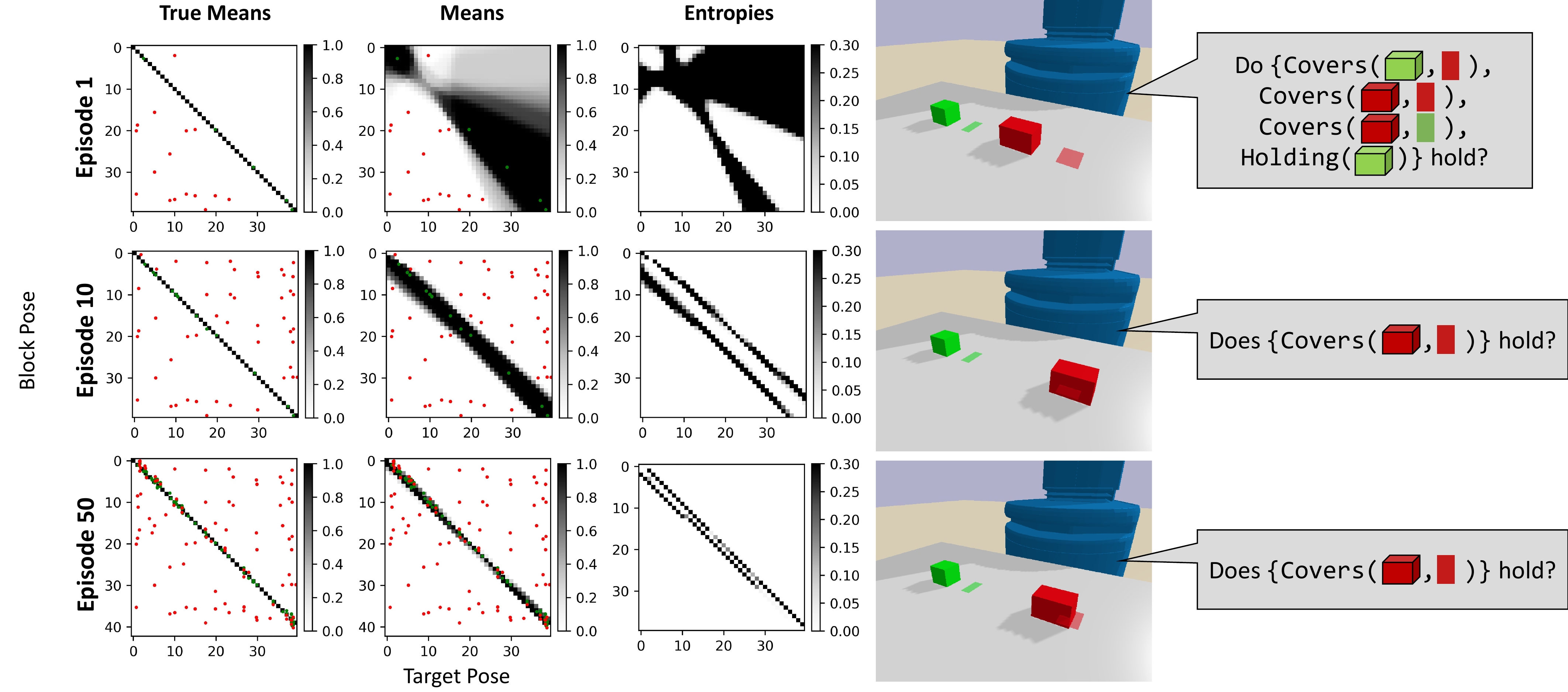}
	\caption{\textbf{Active predicate learning example for a single seed in PickPlace1D.} (Left) The left column (True Means) shows the ground-truth interpretation of the \texttt{Covers} predicate, which holds true (black) when the difference between the target pose and block pose is less than a small threshold. The middle column (Means) shows the agent's predictions averaged over the ensemble, and the right column (Entropies) shows the entropies. Red dots are negative examples and green dots are positive examples. As exploration progresses (top to bottom), the agent makes queries in high-entropy regions and learns better interpretations. (Right) The queries become smaller and more targeted over time.}
  \label{fig:exploration_over_time}
\end{figure}

Our main results are shown in Figure \ref{fig:main_results}.
Comparing the Main approach to \textbf{{\color{black} the action selection baselines}}, we first see that the number of evaluation tasks solved quickly exceeds that of the {\color{black} No Action} baseline.
This confirms that there is value in exploring beyond initial states and that this embodied active learning setting is meaningfully different from standard active learning.
The Main approach is also far more sample-efficient than {\color{black} Random Actions}, supporting the hypothesis that directed exploration is important for active predicate learning.
Finally, we see that Main outperforms {\color{black} GLIB} in PickPlace1D and Blocks and performs similarly in Two Rooms.
These results suggest that exploring via planning to reach specific low-level states, like the partial overlaps in PickPlace1D discussed in Figure~\ref{fig:example_states}, can lead to efficiency gains versus exploring only in the abstract space.
Furthermore, the main lookahead action selection strategy is benefiting from its direct connection to query generation: it considers predicate classifier entropy and does so for every state in the trajectory.
Nonetheless, we believe that GLIB is better able to target goals that are far from the current state than the lookahead action strategy, which relies on random forward sampling.
This may explain GLIB's strong performance in Two Rooms, where the agent should take multiple actions to move from one room to another during exploration.
Combining the strengths of GLIB and lookahead is an exciting direction for future work.

We next compare the Main approach to \textbf{{\color{black} the query generation baselines}}.
As expected, the {\color{black} Ask None} approach performs very poorly because the initial dataset does not contain sufficient class labels to learn good predicate interpretations.
The {\color{black} Ask All} approach performs similarly to Main in terms of evaluation tasks solved but much worse in terms of cumulative query cost.
Ask All continues to accumulate enormous query costs throughout exploration; Main generates a modest number of queries in the beginning, when the agent's uncertainty is high, before plateauing to near ``silence'' when its uncertainty is low.
This confirms that the Main approach is querying enough to learn effectively while avoiding unnecessary queries (e.g., burdening a human expert).
Interestingly, in PickPlace1D, Main seems to slightly outperform Ask All in terms of evaluation tasks solved, though the confidence intervals overlap slightly.
Inspecting the data collected by Ask All, we find a high density of points in regions of input space that are far from the boundary between positive and negative classification; intuitively, we believe these points ``distract'' training from the edge cases that Main is able to ``focus'' on, given its more targeted dataset.
Finally, comparison to the {\color{black} Ask Randomly} approach shows that the Main approach is selecting its queries judiciously.

\footnotetext{The sample size is $n=10$ (the number of random seeds).}

Figure \ref{fig:exploration_over_time} illustrates the Main approach in the PickPlace1D environment.
On the left, we see that entropy for the \texttt{Covers} classifier is initially high in large regions of the input space.
After 10 episodes of exploration, the entropy is much more concentrated around the diagonal of the input space, where the block is partially overlapping the target.
By episode 50, the agent has repeatedly explored states with partial overlaps and refined its classifier further.
On the right, we see that query generation becomes more focused as the classifiers improve.
By episode 50, the agent queries almost exclusively about \texttt{Covers} in states with partial overlap, and evaluation performance is nearly perfect.

In Table~\ref{tab:query}, we analyze the number of queries asked per predicate and find that more difficult predicates are asked about more often.
For example, the \texttt{Covers} predicate dominates the query cost in PickPlace1D, and the \texttt{On} predicate, which requires learning a function that relates the 3D poses of two blocks, is queried about most in Two Rooms and Blocks.

Appendix~\ref{app:experiment-results} reports additional experimental findings.
When we ablate away the MLP ensemble used for modeling predicate classifiers, performance degrades, confirming the importance of modeling epistemic uncertainty.
When we inspect failures to solve evaluation tasks, we see two kinds: failure to find a plan within the timeout, and failure to achieve the goal even when a plan is found due to incorrect goal predicate interpretations.
Finally, we analyze the \texttt{Covers} predicate from PickPlace1D on four illustrative classification examples.

\begin{table*}[t]
    
	\resizebox{\textwidth}{!}{%
    \begin{tabular}{| p{0.75cm} | p{0.75cm} | p{0.75cm} || p{0.75cm} | p{0.75cm} | p{0.75cm} | p{0.75cm} | p{0.7cm} | p{0.7cm} | p{0.9cm} | p{0.7cm} | p{0.7cm} || p{0.7cm} | p{0.7cm} | p{0.9cm} | p{0.7cm} | p{0.7cm} |}
	\hline
        \multicolumn{3}{|c||}{\bf{PickPlace1D}} &
	\multicolumn{9}{c||}{\bf{Two Rooms}} &
	\multicolumn{5}{c|}{\bf{Blocks}} \\
	\hline
    {Hand} & {Cover} & {Hold}&
	{NDial} & {Open} & {On} & {Hold} & {NTab} & {LOff} &{OnTab} & {Clear} & {LOn}& 
	{Hold} & {Open} & {OnTab} & {Clear} & {On}\\
	\hline
     5.5\% & 84.4\% & 10.1\% &
        5.0\% & 5.3\% & 44.8\% & 13.9\% & 4.1\% & 1.7\%  & 11.5\% & 12.0\% & 1.7\% &
        3.8\% & 0.1\% & 12.5\% & 3.4\% & 80.2\% \\\hline
	\end{tabular}}
    
	\caption{\textbf{Query percentages per predicate for Main approach.} Table entries are means over 10 seeds. The predicates from left to right are \texttt{HandEmpty}, \texttt{Covers}, \texttt{Holding},
	\texttt{NextToDial}, \texttt{GripperOpen}, \texttt{On}, \texttt{Holding}, \texttt{NextToTable}, \texttt{LightOff}, \texttt{OnTable}, \texttt{Clear}, \texttt{LightOn}, \texttt{Holding}, \texttt{GripperOpen}, \texttt{OnTable}, \texttt{Clear}, \texttt{On}.
    Predicates with more difficult interpretations are generally included in more queries. For example, in Two Rooms, the \texttt{On} predicate is queried the most and the \texttt{LightOn} predicate is queried the least. Interpreting \texttt{On} requires relating the 3D poses of two blocks, while interpreting \texttt{LightOn} only requires a threshold on a single feature of the light.
 }
	\label{tab:query}
\end{table*}

\section{Related Work}

We now discuss connections to prior work.

\paragraph{Learning state abstractions for TAMP.}
This work contributes to the literature on learning state abstractions for TAMP and decision making more broadly~\citep{li2006towards,Jetchev_learninggrounded,abel2016near,konidaris2018symbols,xu2020daf,akakzia2021grounding,wang2021learning,ahmetoglu2022deepsym,migimatsu2022grounding}.
Particularly relevant is recent work by~\citet{silver2023inventing} who consider learning predicates within their bilevel planning framework~\citep{silver2021learning,chitnis2021learning,silver2022learning}.
Our work is different and complementary in several ways: we learn the interpretations of \emph{known predicates} from \emph{interaction} with an expert in an \emph{online setting}; they learn \emph{latent predicates} from \emph{demonstrations} in an \emph{offline setting}.
Furthermore, they make two key restrictions: predicate classifiers are implemented as simple programs~\citep{pasula2007learning}, and a small set of ``goal predicates'' are given.
Since we have supervision for predicate learning, we are able to instead learn neural network predicate classifiers without given goal predicates.
A straightforward combination of the two would use our approach to learn a small set of predicates from interaction, and their approach to invent additional predicates to aid in planning.

\paragraph{Exploration in relational domains.}
Our action selection strategy takes inspiration from previous work on exploration in relational domains~\citep{walsh2010efficient,rodrigues2011active,ng2019incremental,chitnis2020glib}.
Our lookahead strategy is most similar to the count-based approach considered by \citet{lang2012exploration}, which in turn is related to the classic $\text{E}^3$ approach in the tabular setting~\citep{kearns2002near}.
Also relevant is work by~\citet{andersen2017active}, who consider exploration in the context of learning symbolic state abstractions.
These prior works typically consider finite action spaces, instead of the infinite action space we have here.
Moreover, they operate in the model-based reinforcement learning (MBRL) setting~\citep{eysenbach2018diversity,pathak2019selfsupervised,colas2019curious}, rather than the embodied active learning setting that we consider.

\paragraph{Active learning to ground natural language.}
At the intersection of natural language processing and robotics~\citep{tellex2020robots}, there is longstanding interest in learning to ground language.
For example, \citet{thomason2017opportunistic} consider (non-embodied) active learning for visually grounding natural language descriptions of objects.
\citet{yang2018visual} study natural language query generation and propose an RL-based approach for selecting informative queries.
\citet{roesler2019action} learn to ground natural language goals and learn policies for achieving those goals.
We differ from these previous works in our focus on learning to plan and planning to learn.
Given recent interest in using large language models (LLMs) for planning~\citep{huang2022language,li2022pre,ahn2022can,sharma2022skill}, a possible direction for future work would combine active learning for natural language grounding and LLM-based planning.
However, recent studies suggest that classical AI planning techniques are still much stronger than LLM-based planners \citep{silver2022pddl,valmeekam2023planning}.

\paragraph{Embodied active learning.}
The challenge of interleaving action selection and active information gathering has been considered from many perspectives including \emph{active reward learning}~\citep{daniel2014active,schulze2018active,krueger2020active}, \emph{active preference learning}~\citep{sadigh2017active,biyik2018batch} and \emph{interactive perception}~\citep{bohg2017interactive,jayaraman2018learning}.
We are especially influenced by \citet{noseworthy2021active}, who actively learn to estimate the feasibility of abstract plans in TAMP.
Also notable is recent work by~\citet{lamanna2023planning}, who use known operators and AI planning methods to learn object properties through online exploration of a robotic environment.
Finally, \citet{kulick2013active} consider active learning for relational symbol grounding, but in a non-sequential and discrete-action setting.
Embodied active learning also shares certain facets with lifelong learning \citep{thrun1998lifelong,abel2018state} in that the agent improves incrementally and accumulates knowledge that helps it become better at learning in the future.
Lifelong learning approaches that make use of hierarchy and abstraction for decision-making are most related to our efforts~\citep{tessler2017deep,wu2020model,lu2020reset}.
However, unlike lifelong learning, we do not address learning in non-stationary environments, nor do we attempt to learn incrementally (we retrain models from scratch).

\section{Conclusion}

In this paper, we proposed an embodied active learning framework for learning predicates useful for TAMP in continuous state and action spaces.
Through experiments, we showed that the predicates are learned with strong sample efficiency in terms of both number of environment transitions and number of queries to the expert.

\paragraph{Limitations and Future Work.} There are limitations of the present work and challenges for active predicate learning in general.
In this work, we assumed an object-centric view of a fully-observed state; access to a deterministic simulator; and access to hybrid controllers.
There is work on removing each of these assumptions with learning~\citep{yuan2021sornet,wang2022generalizable,chitnis2021learning,silver2022learning}, but integration would be nontrivial.
Assuming access to a deterministic simulator that exactly matches the environment is particularly unrealistic in real-world settings. In Appendix~\ref{app:experiment-results}, we present additional results with a noisy (but still known) simulator.
We also used noise-free scripts to generate the initial demonstrations and expert responses.
Since the agent is primarily learning from its own experience, we expect some robustness to noise in the initial demonstrations.
To handle noise in the expert responses, we could model aleatoric uncertainty in addition to epistemic uncertainty, perhaps through the BALD objective~\citep{houlsby2011bayesian,noseworthy2021active}.
In Appendix~\ref{app:experiment-results}, we also present additional experimental results where the expert gives noisy predicate labels.

For active predicate learning in general, one challenge is determining how frequently to relearn models.
We relearned models after every episode, which led to strong sample complexity, but slowed down experiments overall (one run typically taking between 3 and 36 hours).
Incremental learning approaches, especially for training the neural network predicate classifiers and samplers, could provide useful speedups~\citep{castro2018end,ng2019incremental}.
Another issue is that the predicates given by the expert may be insufficient, or even unhelpful, for TAMP.
Combining our approach with that of~\citet{silver2023inventing} would help to address this issue since the agent could invent its own predicates, but we should also allow the agent to drop or modify expert-given predicates that it deems unhelpful.
Finally, if learning predicates is one component of a larger learning-to-plan system, then predicate classifier entropy should not be the only driver of action selection: the agent's desire to learn better operators, samplers, controllers, state features, and so on, should also play a role in exploration.

\section{Acknowledgements}

We gratefully acknowledge support from NSF grant 2214177; from AFOSR grant FA9550-22-1-0249; from ONR
MURI grant N00014-22-1-2740; from ARO grant W911NF-23-1-0034; from the MIT-IBM Watson Lab; from the
MIT Quest for Intelligence; and from the Boston Dynamics Artificial Intelligence Institute.
Tom is supported by a NSF Graduate Research Fellowship.
We thank Jorge Mendez, Rohan Chitnis, Willie McClinton, and Leslie Kaelbling for helpful comments on an earlier draft.

\bibliography{collas2023_conference}
\bibliographystyle{collas2023_conference}

\clearpage

\appendix
\section{Appendix}

\subsection{Additional Experimental Details}
\label{app:experiment-details}

We now provide additional details for reproducing our experimental results.

\subsection{Additional Details for Approaches}

\paragraph{Learning.} 
All neural networks are trained with the Adam optimizer \citep{kingma2014adam}.
For the classifier networks, each ensemble consists of $10$ MLPs, and every MLP is trained for 100K epochs.
Each MLP consists of two hidden layers, each of size 32, with ReLU activations.
As mentioned in the main text, we use custom weight initialization to facilitate diversity in the ensemble.
In particular, we initialize the MLP weights according to a $\mathcal{N}(0, 1)$ distribution.
If the model fails to converge during training, we try reinitializing and retraining the model up to five times in total.
Sampler neural network architecture and training is identical to that of~\citet{chitnis2021learning}, except that we forgo training a discriminator and rejection sampling for simplicity.
Operator learning is also identical to the prior work, except that we filter out any operators with less than $10$ data points to facilitate efficient learning and planning, which was not needed in prior work because operators were learned from demonstrations instead of exploration data.

\paragraph{Planning.} 
For planning in the evaluation tasks, all experiments use $\text{A}^{*}$ search with the LMCut heuristic for abstract planning, with the implementation for LMCut taken from Pyperplan~\citep{pyperplan}.
Following previous work, we consider up to $n_{\text{abstract}} = 8$ abstract plans per task and $n_{\text{samples}} = 10$ samples per step during refinement.
Planning is timed out after 10 seconds.

\paragraph{Approach-specific details.}
The main entropy-based query policy uses a threshold of $\alpha = 0.05$, which we tuned to optimize performance in PickPlace1D.
The GLIB baseline is GLIB-L1 from~\citet{chitnis2020glib}.
The Ask Randomly baseline asks about each possible ground atom with 0.03 probability, which we selected to approximately match the query rate of Main.

\subsection{Additional Details for Environments}

\begin{figure}[h]
    \centering
	\includegraphics[width=\textwidth]{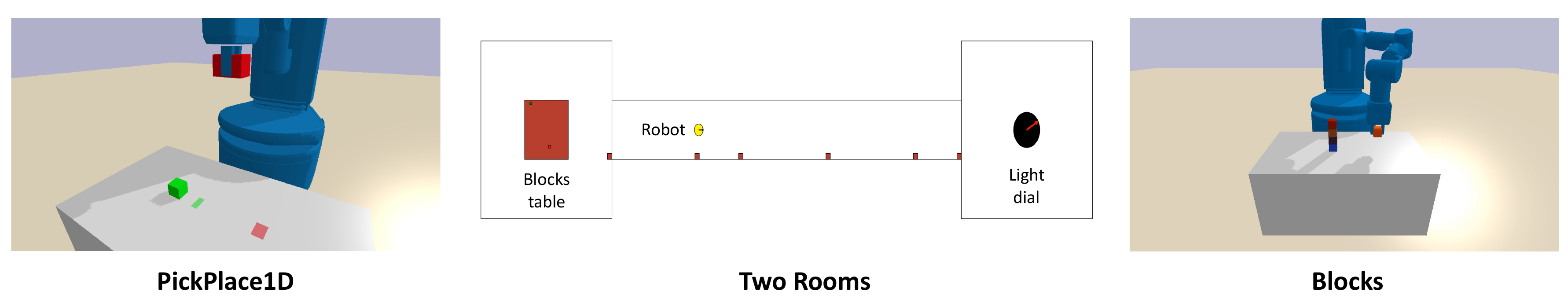}
	\caption{\textbf{Environments.} PickPlace1D and Blocks are from~\citet{chitnis2021learning}, and Two Rooms is original to this work but loosely inspired by~\citet{konidaris2009efficient}.}
  \label{fig:environments}
\end{figure}

See Figure~\ref{fig:environments} for renderings of each environment.
The details below for PickPlace1D and Blocks are modified with permission from \citet{silver2023inventing}.

\paragraph{PickPlace1D.}
In this environment, a robot must pick blocks and place them onto target regions along a table surface.
All pick and place poses are on a 1D line. The three object types are block, target, and robot. Blocks and targets each have two features for their pose and width.
Robots have three features: a 2D pose and the (symmetric) value of the finger joint.
The block widths are larger than the target widths, and the goal requires each block to be placed so that it completely covers the respective target region.
The predicates to learn include \texttt{Covers(?block, ?target)}, \texttt{Holding(?robot, ?block)}, and \texttt{HandEmpty(?robot)}.
There is only one controller, \texttt{PickPlace}, with no discrete parameters; its continuous parameter is a single real number denoting the location to perform either a pick or a place, depending on the current state of the robot's gripper.
Each action updates the state of at most one block, based on whether any is in a small radius from the continuous parameter.
During exploration and evaluation there are 2 blocks, 2 targets, and 1 robot.
In each task, with 75\% probability the robot starts out holding a random block; otherwise, both blocks start out on the table.
This environment was established by~\citet{silver2021learning}, but that work involved manually defined state abstractions, which we do not provide in this paper.

\paragraph{Two Rooms.}
In this environment, a robot in 3D starts out in a room with 3 blocks on a table.
Beyond this room lies a hallway, and on the other end of the hallway is another room with a dial that controls a light.
The three object types are block, robot, and dial.
Blocks have five features: a 3D pose, a bit for whether it is currently grasped, and a bit for whether there the block is clear from above.
The latter feature was added to make learning the \texttt{Clear(?block)} predicate possible, since we restrict our predicate classifiers to consider only the states of its arguments (Section~\ref{sec:methods}).
The robot has four features: a 2D position  rotation for the base, and a (symmetric) value for the finger joints.
The dial has three features: a 2D position and a level, where the level controls the light.
The goal of an evaluation task is to build a certain block tower and turn the light on or off.
There are six controllers: \texttt{Pick}, \texttt{Stack}, \texttt{PutOnTable}, \texttt{MoveTableToDial}, \texttt{TurnOnDial}, and \texttt{TurnOffDial}.
The first three controllers are identical to Blocks; see below.
The last three controllers are each parameterized by a robot, dial, and a 3D continuous change in pose.
The predicates to learn include \texttt{On(?block1, ?block2)}, \texttt{OnTable(?block)}, \texttt{GripperOpen(?robot)}, \texttt{Holding(?robot, ?block)}, \texttt{Clear(?block)}, \texttt{NextToTable(?robot)}, \texttt{NextToDial(?robot, ?dial)}, \texttt{LightOn(?dial)}, and
\texttt{LightOff(?dial)}.
There are 3 blocks during exploration and in the evaluation tasks.
This environment is original to this work.

\paragraph{Blocks.}
In this environment, a robot in 3D must interact with blocks on a table to assemble them into towers.
This is a robotics adaptation of the blocks world domain in AI planning.
The two object types are block and robot. Blocks have five features: an x/y/z pose, a bit for whether it is currently grasped, and a bit for whether there the block is clear from above.
The latter feature was added to make learning the \texttt{Clear(?block)} predicate possible, since we restrict our predicate classifiers to consider only the states of its arguments (Section~\ref{sec:methods}).
Robots have four features: x/y/z end effector pose and a (symmetric) value for the finger joints.
There are three controllers: \texttt{Pick}, \texttt{Stack}, and \texttt{PutOnTable}. \texttt{Pick} is parameterized by a robot and a block to pick up. \texttt{Stack} is parameterized by a robot and a block to stack the currently held one onto.
\texttt{PutOnTable} is parameterized by a robot and a 2D place pose representing normalized coordinates on the table surface at which to place the currently held block.
The predicates to learn include \texttt{On(?block1, ?block2)}, \texttt{OnTable(?block)}, \texttt{GripperOpen(?robot)}, \texttt{Holding(?robot, ?block)}, and \texttt{Clear(?block)}.
During exploration there is 3 or 4 blocks, while evaluation tasks involve 5 or 6 blocks.
This environment was established by~\citet{silver2021learning}, but that work involved manually defined state abstractions, which we do not provide in this paper.

\subsection{Additional Experimental Results}
\label{app:experiment-results}

Here we present additional experimental findings.

\begin{figure}[h]
    \centering
	\includegraphics[width=0.9\textwidth]{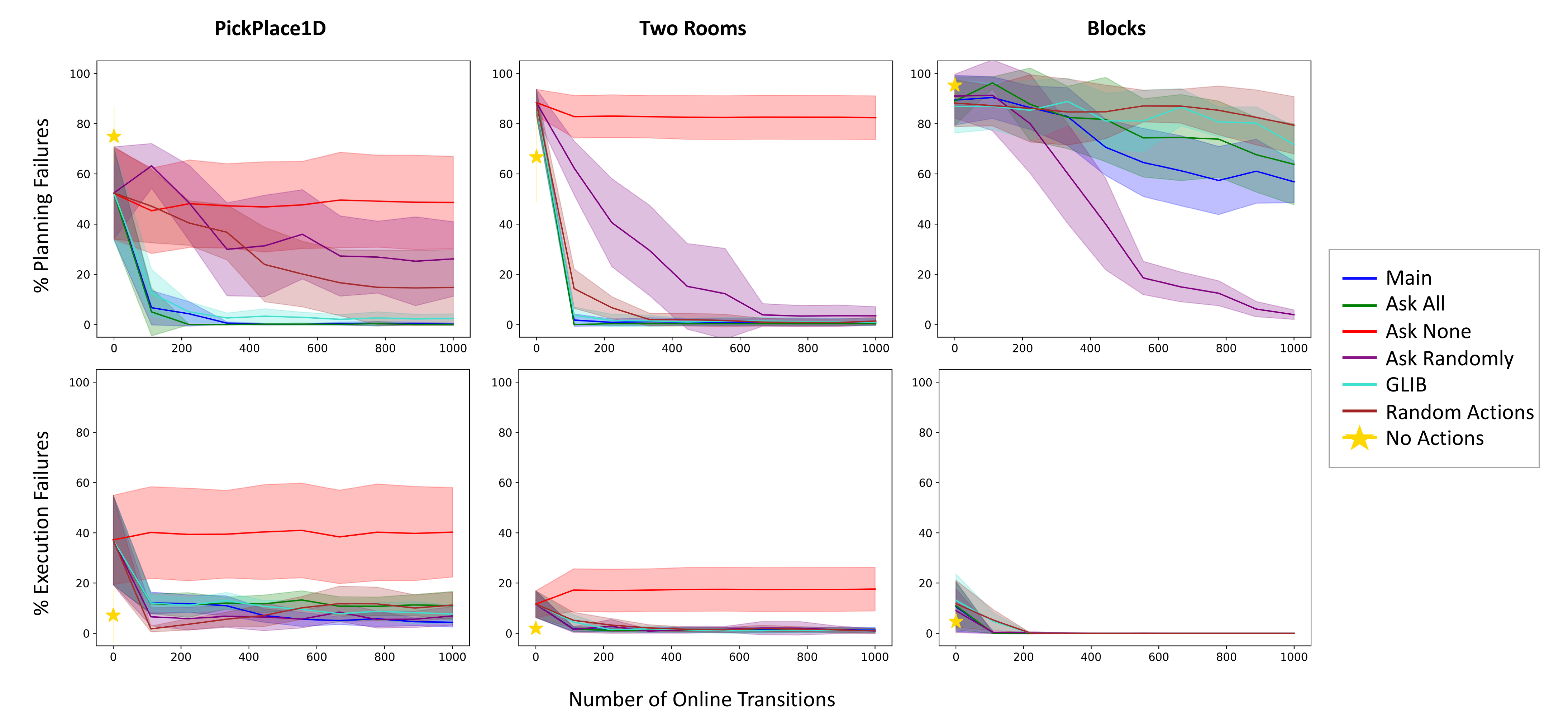}
	\caption{\textbf{Failure analysis.} All results are averaged over 10 random seeds. Lines are means and shaded regions are 95\% $t$-confidence intervals. See text for details.}
  \label{fig:failure_results}
\end{figure}

\paragraph{Failure analysis.}
Figure~\ref{fig:failure_results} decomposes evaluation task failures into two categories: planning failures and execution failures.
Planning failures occur when the agent cannot find a plan within the 10 second timeout and can be due to poor predicate interpretations, poor operators, or poor samplers.
Execution failures occur when the agent finds a plan but fails to reach the goal upon executing the plan in the environment.
Since the agent has a perfect simulator of the deterministic environment, this type of failure only occurs when the agent has an incorrect interpretation of a goal predicate.
For example, while planning with the simulator, the agent may find a plan that it believes achieves the goal \texttt{Covers(block1, target)}, but when it executes the plan, it actually fails to cover \texttt{block1} with \texttt{target} according to the expert's (ground-truth) interpretation of \texttt{Covers}.

\begin{figure}[h]
    \centering
	\includegraphics[width=0.9\textwidth]{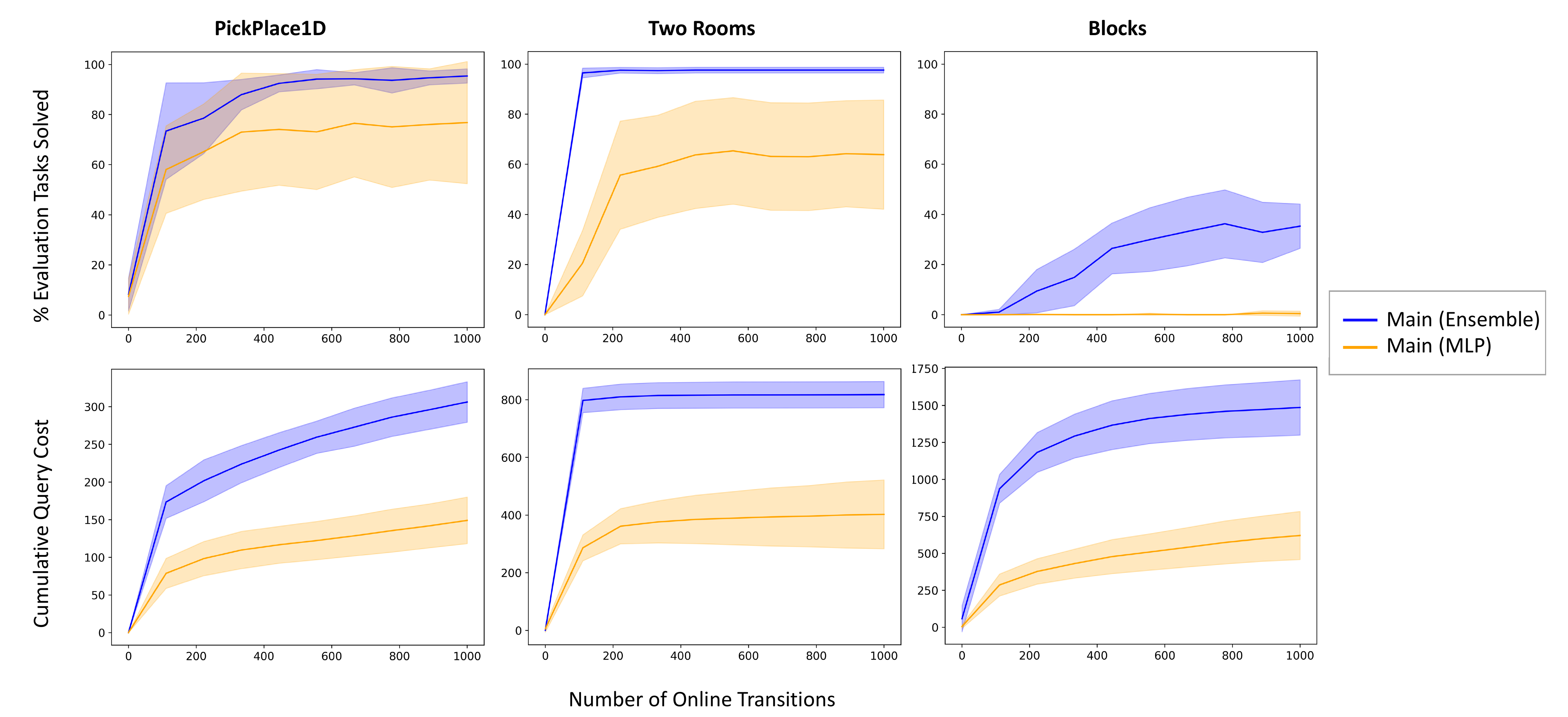}
	\caption{\textbf{Ensemble ablation.} All results are averaged over 10 random seeds. Lines are means and shaded regions are 95\% $t$-confidence intervals. See text for details.}
  \label{fig:mlp_vs_ensemble}
\end{figure}

\paragraph{Ensemble ablation.}
Figure~\ref{fig:mlp_vs_ensemble} compares our main approach to an ablation that uses a single neural network for each learned predicate interpretation instead of an ensemble.
Entropy is calculated on the basis of that single MLP neural network's predicted class probability.
The stark difference between the ensemble and the single MLP confirms that an ensemble is vital for representing uncertainty in all environments.

\begin{figure}[h]
    \centering
	\includegraphics[width=0.9\textwidth]{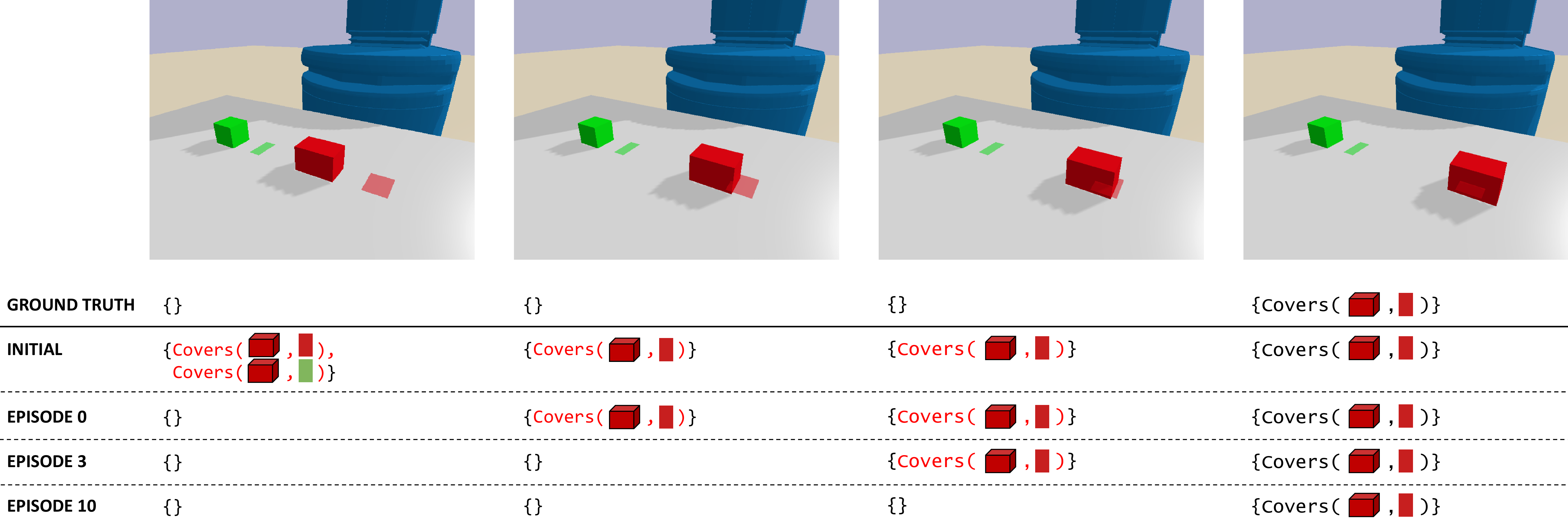}
	\caption{\textbf{PickPlace1D Held-out Test Cases.} From a single representative seed. Red text indicates a false interpretation compared to ground-truth while black indicates true. See text for details.}
  \label{fig:pickplace1d_heldout_examples}
\end{figure}

\paragraph{PickPlace1D Held-out Test Cases.}
In Figure~\ref{fig:pickplace1d_heldout_examples}, we show four illustrative states from PickPlace1D to demonstrate how the learned interpretation of the \texttt{Covers} predicate changes as the robot explores under the entropy-based query policy.
Specifically, each set in the figure contains the ground atoms that result from the robot applying its interpretation of \texttt{Covers} to one of the four states at some point in time during learning.
Ground-truth atoms are shown in the top row.
Intuitively, the first (leftmost) and fourth (rightmost) states have the easiest classification problems for \texttt{Covers} because in the first, the green and red block are both clearly far from covering the target regions, and in the fourth, the red block completely overlaps the red target region.
From just one episode after initialization, the agent correctly learns the classification for and becomes certain about these two states.
The second and third states offer progressively more difficult classification problems, requiring more exploration.
The agent classifies the second problem correctly by episode 3, and the third problem by episode 10.

\begin{figure}[h]
    \centering
	\includegraphics[width=0.9\textwidth]{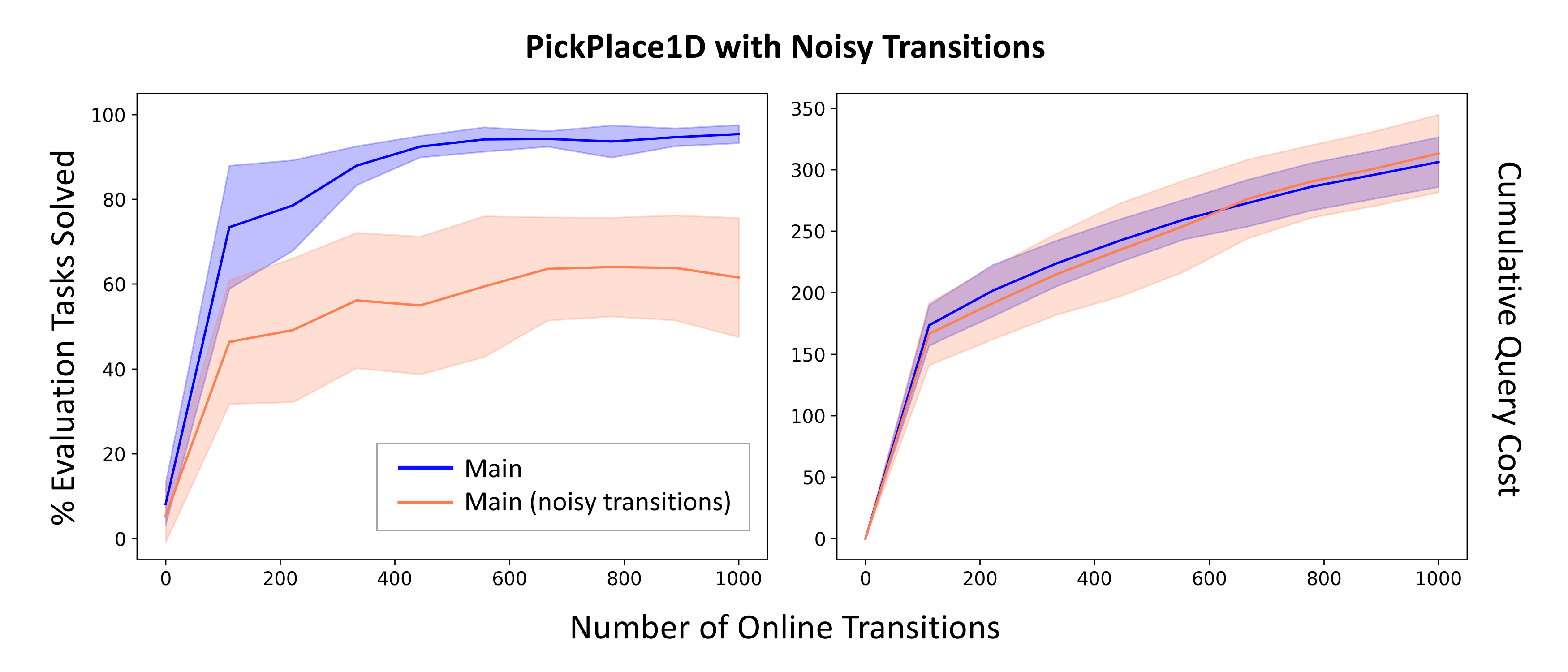}
	\caption{\textbf{PickPlace1D with Noisy Transitions.} All results are averaged over 10 random seeds. Lines are means and shaded regions are 95\% $t$-confidence intervals. See text for details.}
  \label{fig:noisy_transitions}
\end{figure}

\paragraph{Impact of Noisy Transitions.}
Figure \ref{fig:noisy_transitions} shows the impact of noisy environment transitions on our main approach.
This experiment uses a version of the PickPlace1D environment such that when the robot takes an action to place a block on a target, the new position of the block is perturbed according to a $\mathcal{N}(0, 0.015)$ distribution.
We find that this noticeably impacts the performance of our main approach, which is expected since the robot may encounter different noise during planning vs. during evaluation.
Interestingly, the average query cost incurred is largely the same as before, suggesting that the robot’s exploration and querying is similar to before, and since the robot fully observes its environment, that would mean the performance hit is largely attributed to planning.

\begin{figure}[h]
    \centering
	\includegraphics[width=0.9\textwidth]{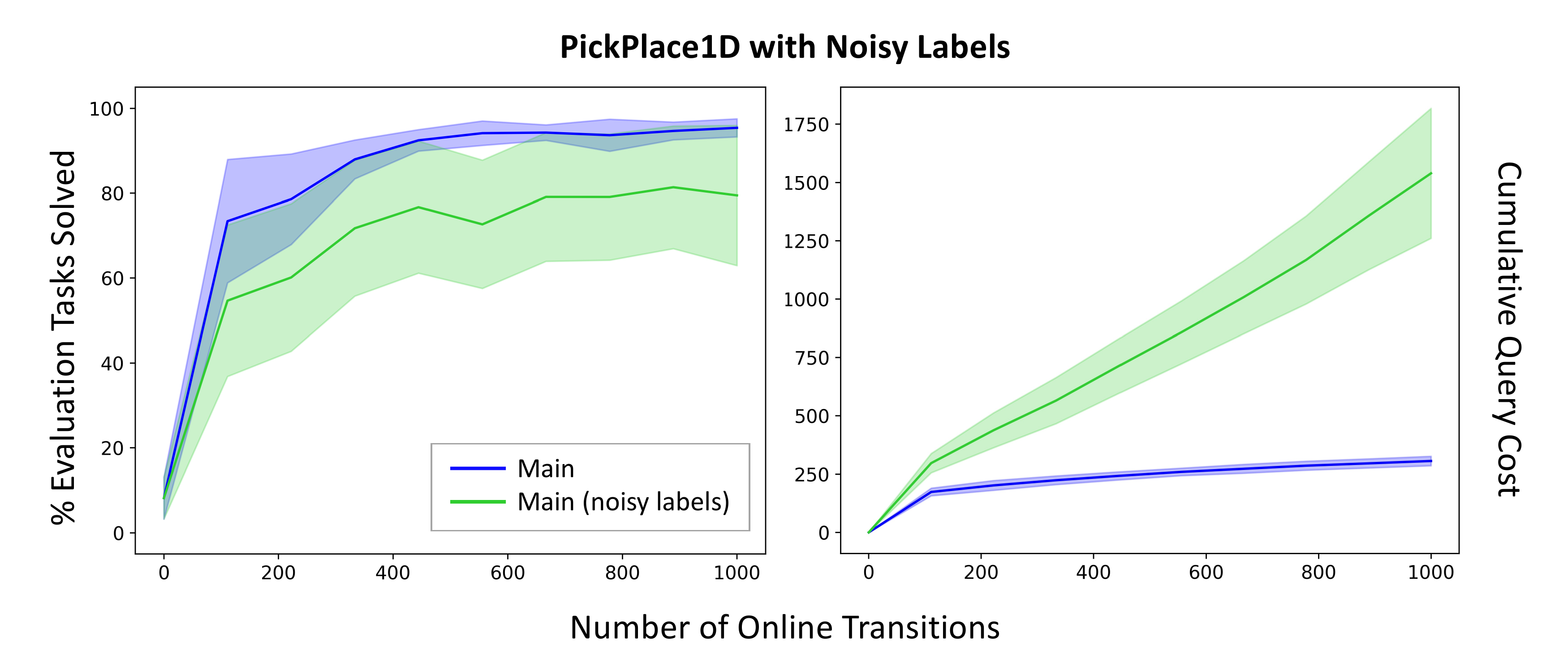}
	\caption{\textbf{PickPlace1D with Noisy Labels.} All results are averaged over 10 random seeds. Lines are means and shaded regions are 95\% $t$-confidence intervals. See text for details.}
  \label{fig:noisy_labels}
\end{figure}

\paragraph{Impact of Noisy Predicate Labels.}
Figure \ref{fig:noisy_labels} shows the impact of noisy predicate labels on our main approach.
In this experiment, any given ground atom label in an expert’s response to a query is randomly inverted (made incorrect) with probability 0.05.
We find that the noisy labels have a noticeable detrimental impact on performance: despite incurring much higher query cost, the robot solves fewer evaluation tasks, and there is greater variation in performance.
We hypothesize that this is largely due to incorrectly labeled data points for the Covers predicate, since we have demonstrated that its classification boundary is tricky to learn (Figure \ref{fig:exploration_over_time}).

\end{document}